\documentclass[10pt,twocolumn,letterpaper]{article}

\usepackage[pagenumbers]{iccv}

\usepackage[accsupp]{axessibility}  

\usepackage{xcolor, colortbl}
\usepackage{amsmath}
\usepackage{amssymb}
\usepackage{algorithm}
\usepackage{multicol}
\usepackage{multirow}

\definecolor{mono_orange}{RGB}{237,125,49}
\definecolor{mono_teal}{RGB}{143,206,211}

\usepackage{soul}
\newcommand{\hlorange}[1]{{\sethlcolor{mono_orange!20}\hl{#1}}}
\newcommand{\hlteal}[1]{{\sethlcolor{mono_teal!30}\hl{#1}}}

\definecolor{iccvblue}{rgb}{0.21,0.49,0.74}
\usepackage[pagebackref=false,breaklinks,colorlinks,allcolors=iccvblue]{hyperref}

\definecolor{car}{rgb}{0.39215686, 0.58823529, 0.96078431}
\definecolor{bicycle}{rgb}{0.39215686, 0.90196078, 0.96078431}
\definecolor{motorcycle}{rgb}{0.11764706, 0.23529412, 0.58823529}
\definecolor{truck}{rgb}{0.31372549, 0.11764706, 0.70588235}
\definecolor{other-vehicle}{rgb}{0.39215686, 0.31372549, 0.98039216}
\definecolor{person}{rgb}{1., 0.11764706, 0.11764706}
\definecolor{bicyclist}{rgb}{1., 0.15686275, 0.78431373}
\definecolor{motorcyclist}{rgb}{0.58823529, 0.11764706, 0.35294118}
\definecolor{road}{rgb}{1., 0., 1.}
\definecolor{parking}{rgb}{1., 0.58823529, 1.}
\definecolor{sidewalk}{rgb}{0.29411765, 0., 0.29411765}
\definecolor{other-ground}{rgb}{0.68627451, 0., 0.29411765}
\definecolor{building}{rgb}{1., 0.78431373, 0.}
\definecolor{fence}{rgb}{1., 0.47058824, 0.19607843}
\definecolor{vegetation}{rgb}{0., 0.68627451, 0.}
\definecolor{trunk}{rgb}{0.52941176, 0.23529412, 0.}
\definecolor{terrain}{rgb}{0.58823529, 0.94117647, 0.31372549}
\definecolor{pole}{rgb}{1., 0.94117647, 0.58823529}
\definecolor{traffic-sign}{rgb}{1., 0., 0.}   

\usepackage{titletoc}
\usepackage[toc,page]{appendix}

\title{Monocular Semantic Scene Completion via  Masked Recurrent Networks}

\author{
Xuzhi Wang$^1$ \quad
Xinran Wu$^1$ \quad
Song Wang$^2$ \quad
Lingdong Kong$^3$\thanks{Corresponding authors.}  \quad
Ziping Zhao$^1$\footnotemark[1] \\
$^1$Tianjin Normal University \quad
$^2$Zhejiang University \quad
$^3$National University of Singapore \\
}

\begin{document}

\maketitle

\begin{abstract}
Monocular Semantic Scene Completion (MSSC) aims to predict the voxel-wise occupancy and semantic category from a single-view RGB image. Existing methods adopt a single-stage framework that aims to simultaneously achieve visible region segmentation and occluded region hallucination, while also being affected by inaccurate depth estimation. Such methods often achieve suboptimal performance, especially in complex scenes. We propose a novel two-stage framework that decomposes MSSC into coarse MSSC followed by the Masked Recurrent Network. Specifically, we propose the Masked Sparse Gated Recurrent Unit (MS-GRU) which concentrates on the occupied regions by the proposed mask updating mechanism, and a sparse GRU design is proposed to reduce the computation cost. Additionally, we propose the distance attention projection to reduce projection errors by assigning different attention scores according to the distance to the observed surface. Experimental results demonstrate that our proposed unified framework, MonoMRN, effectively supports both indoor and outdoor scenes and achieves state-of-the-art performance on the NYUv2 and SemanticKITTI datasets. Furthermore, we conduct robustness analysis under various disturbances, highlighting the role of the Masked Recurrent Network in enhancing the model's resilience to such challenges. The source code is publicly available at: \url{https://github.com/alanWXZ/MonoMRN}.
\end{abstract}    
\section{Introduction}
\label{sec:intro}

Semantic scene completion is an essential task in 3D scene understanding, which aims to jointly infer the holistic semantics and geometry of the 3D scene from partial observation \cite{geiger2012kitti,cao2021pcam,martyniuk2025lidpm,li2022coarse3d}. The task is motivated by the fact that the occupancy patterns and semantic labels of the scene are mutually reinforcing~\cite{SSCNet,bian2025dynamiccity,lin2022sparse4d,li2025seeground,cao2023scenerf}. Researchers gradually discover the important applications of semantic scene completion in indoor scene analysis, robotics, virtual reality, and many downstream tasks in autonomous driving.

Recently, the completion of monocular 3D semantic scenes has become a hot topic of research. Compared with methods relying on 3D sensors like LiDAR or depth sensors and stereo vision using RGB stereo cameras, monocular-based approaches offer advantages such as lower cost, simpler configuration, and greater portability \cite{ssc_outdoor_MonoScene,liu2023seal,xu2024superflow,kong2025largead,nunes2023tarl}. Additionally, compared with stereo and surround-view setups, a monocular configuration offers greater adaptability to both indoor and outdoor environments, as indoor scenarios particularly emphasize portability. A monocular setup facilitates seamless cross-scenario generalization, making it a more versatile choice for a wide range of applications.

\begin{figure}
    \centering
    \includegraphics[width=\linewidth]{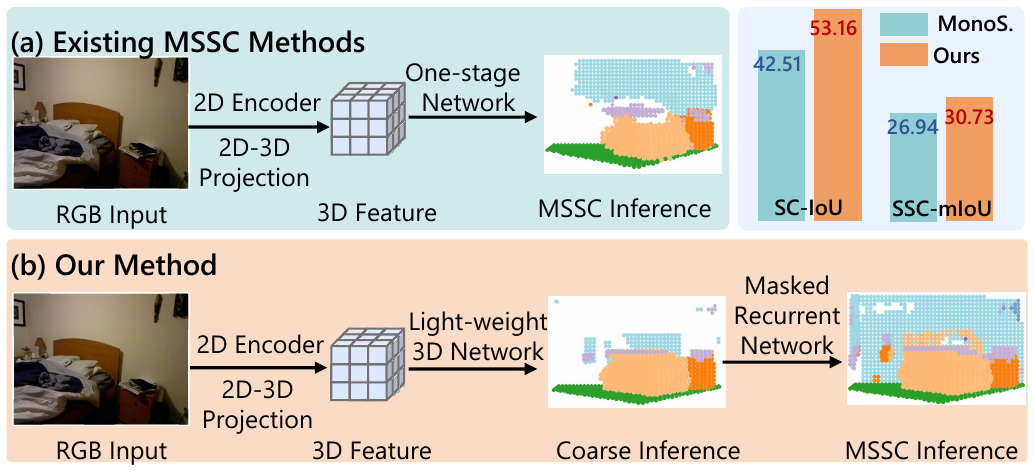}
    \vspace{-0.6cm}
    \caption{The framework of our method and existing methods. Our method decomposes MSSC into coarse SSC and Masked Recurrent Network. MonoS. represents the work of MonoScene~\cite{ssc_outdoor_MonoScene}.}
    \label{fig:teaser}
    \vspace{-4mm}
\end{figure}

Most existing Monocular Semantic Scene Completion (MSSC) models \cite{ssc_outdoor_MonoScene,ssc_outdoor_VoxFomer} attempt to simultaneously complete the coupled 3D segmentation and completion tasks in a single stage. However, this framework presents significant challenges, particularly in complex scenes, where performance tends to be poor. When performing scene completion, it is necessary not only to restore the geometric structure but also to ensure the consistency of semantic labels. At the same time, during scene segmentation, the accuracy and completeness of the completion results directly affect the quality of the segmentation \cite{ssc_outdoor_MonoOcc,ssc_outdoor_MonoScene,chen2023clip2Scene,chen2023towards,sautier2025clustering,puy2023waffle}. Moreover, most existing methods~\cite{ssc_outdoor_VoxFomer} rely on estimated depth maps to provide geometric information. The noise introduced by inaccurate depth estimations further intensifies the complexity and challenges associated with the problem.

In contrast to adopting a one-stage framework, it is natural to decompose MSSC into several stages. Therefore, we propose a new two-stage framework, named MonoMRN, that decomposes MSSC into initial estimation and a novel Masked Recurrent Network for iterative refinement. The GRU (Gated Recurrent Unit) is also used to update partial features based on the previous stage, aiming to reduce task complexity and improve performance.

Due to the standard GRU processing of all positions of the input feature, it results in significant computational overhead. We propose the MS-GRU (Masked Sparse Gated Recurrent Unit), which enhances feature sparsity by applying a dynamic mask that focuses on the occupancy region and designs a sparse version of the GRU to leverage this sparsity for acceleration. Moreover, motivated by the fact that completion and segmentation tasks mutually reinforce each other, we designed a dynamic mask module, which updates the mask at each iteration stage. Additionally, to efficiently update the MS-GRU in each iteration, we propose the distance attention projection, which effectively reduces projection error by reasonably distributing feature weights.

\begin{figure*}
    \centering
    \includegraphics[width=\linewidth]{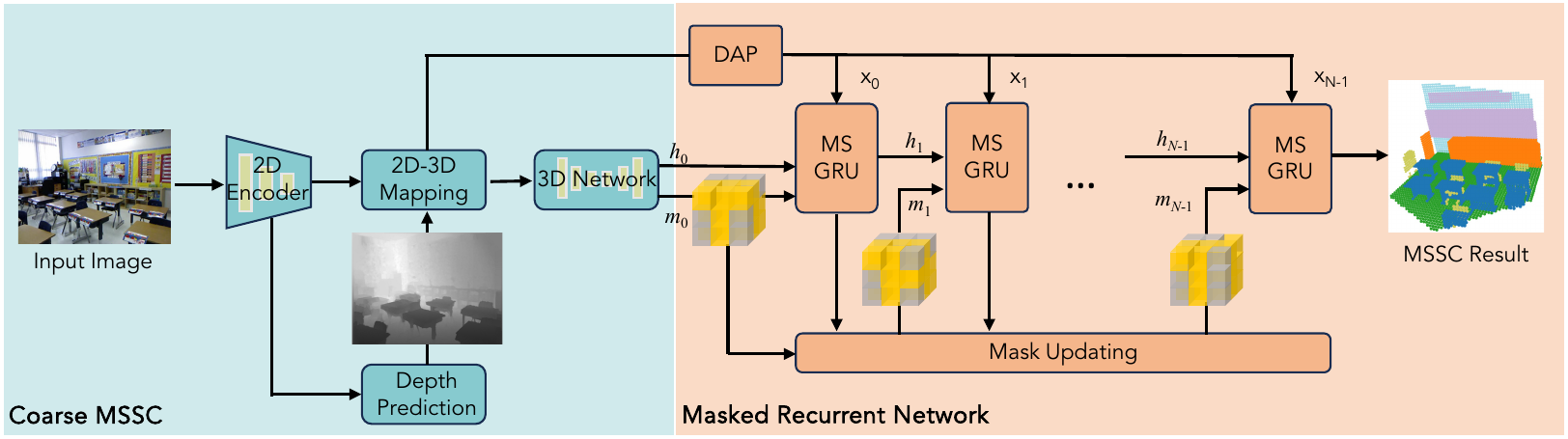}
    \caption{The overall framework of MonoMRN. We first employ the base network to produce the coarse MSSC $h_0$ and the mask $m_0$. Subsequently, a series of MS-GRUs operate in a recurrent manner to update MSSC estimation. For each iteration, the mask is updated by the proposed Mask Updating Module. `DAP' represents the Distance Attention Projection.}
    \label{fig:framework}
\end{figure*}

The overall framework is shown in \cref{fig:framework}. Our method consists of two stages: Coarse MSSC and Masked Recurrent Network. For the initial estimation stage, the single RGB input is fed into the encoder to extract 2D features. The 2D features are projected to 3D space according to the estimated depth. Subsequently, the projected features are fed into the 3D network to obtain a coarse MSSC prediction. For the recurrent refinement stage, the MS-GRU module takes the distance attention projection $x_0$, initial estimation feature $h_0$, and initial mask $m_0$ as input to recurrently update the semantic scene completion results. The mask is updated in each iteration by the mask updating module.

The key contributions can be summarized as follows: 
\begin{itemize}
    \item We propose a novel semantic scene completion framework, MonoMRN, that decomposes semantic scene completion to coarse MSSC and Masked Recurrent Network for iterative refinement.

    \item We design a spectrum of useful components in our framework. The MS-GRU is proposed to focus on the occupied region to enhance performance, while also accelerating the process by increasing sparsity. We propose a mask update mechanism to update the mask during each iteration. We propose the distance attention projection to reduce the projection error. 

    \item Extensive experiments validate the effectiveness of our proposed approach. We achieve state-of-the-art performance on the indoor NYUv2 and outdoor SemanticKITTI datasets. We further validate the robustness against various disturbances and reveal the role of the Masked Recurrent Network in the network optimization process.
\end{itemize}
\section{Related Work}

\noindent\textbf{3D from a Single Image.}
The exploration of estimating object or scene geometry from a single-view RGB image has been a longstanding area of study. Early researchers focus on single/multi 3D objects reconstruction by explicit representation \cite{3D_RGB_2018icml,3D_RGB_3D-R2N2,3D_RGB_point_set,LiDAR_multi-modal-pami,hong20244ddsnet,boulch2023also,puy2024scalr,samet2023seeding,michele2024train} or implicit representation \cite{3D_RGB_DeepSDF,3D_RGB_convolution_occupancy,3D_RGB_occupancy_network,xu2025frnet,wang2025pointlora,choy2019minkunet,tang2020spvcnn,li2023lim3d,sautier2024bevcontrast,li2024rapid-seg,samet2024milan}. Recently, some researchers have focused on scene-level 3D reconstruction from a single image. The pioneer work~\cite{3D_RGB_panoptic} unifies 3D reconstruction, 3D semantic segmentation, and 3D instance segmentation into a task known as panoptic 3D scene reconstruction \cite{hong20244ddsnet}. BUOL~\cite{3D_RGB_BUOL} proposes a bottom-up framework with occupancy-aware lifting to address instance-channel ambiguity and voxel-reconstruction ambiguity. Recently, there has been a surge of interest in monocular semantic scene completion. The seminal work MonoScene \cite{ssc_outdoor_MonoScene} conducts monocular semantic scene completion by line of sight projection and proposes a 3D context relation prior to enforce spatial-semantic consistency.

\noindent\textbf{Semantic Scene Completion from 3D.}
The initial SSC methods \cite{SSCNet,ESSCNet} rely on depth data and are designed for indoor scenarios. To further enhance the performance of SSC, researchers \cite{FFNet,SATNet,SISNet,SATNet,SATNet} have explored the integration of complementary RGB-D data, leveraging both geometric depth information and rich color cues from RGB images. In the past few years, outdoor SSC ~\cite{ssc_outdoor_LMSCNet,ssc_outdoor_S3CNet,ssc_outdoor_JS3CNet,ssc_outdoor_SCPNet} has gained much attention because it can aid in holistic scene understanding. JS3CNet~\cite{ssc_outdoor_JS3CNet} explores and enhances the interaction between LiDAR segmentation \cite{LiDAR_LaserMix,LiDAR_range,LiDAR_adapt_sam,LiDAR_multi-modal-pami,LiDAR_multi-space,LiDAR_LiDAR2Map,LiDAR_Meta-rangeseg,LiDAR_LiMoE,LiDAR_NUC-Net,liu2023uniseg,cheng2021af2s3net,yan20222dpass,ando2023rangevit}, and Semantic Scene Completion (SSC) through a point-voxel interaction module. SCPNet~\cite{ssc_outdoor_SCPNet} redesigns the completion sub-network using dense-to-sparse knowledge distillation.

\noindent\textbf{Semantic Scene Completion from RGB.}
The RGB-based SSC can be categorized into two types based on application scenarios: 1) unified methods~\cite{ssc_outdoor_MonoScene,ssc_NDC-scene} applicable to both indoor and outdoor scenes, and 2) methods specifically designed for outdoor scenes~\cite{ssc_outdoor_Symphonize,BRGScene,ssc_outdoor_OccFormer,ssc_outdoor_VoxFomer,ssc_outdoor_HASSC,ssc_outdoor_Bi-SSC,ssc_outdoor_H2gformer,ssc_outdoor_TPVFormer,sautier2022slidr,mahmoud2023st-slidr,chen2024csc}. 
The first category is unified methods for both indoor and outdoor scenes. MonoScene~\cite{ssc_outdoor_MonoScene} first proposes a monocular semantic scene completion method, which is a unified method for both indoor and outdoor scenes. It introduces a line of sight projection and a 3D context relation prior to enforcing spatio-semantic consistency. NDCScene~\cite{ssc_NDC-scene} identifies several critical issues in monocular semantic scene completion, including the feature ambiguity of size, depth, and pose.
The second category focuses on outdoor scenes \cite{geiger2012kitti,dataset_semantickitti}. VoxFormer~\cite{ssc_outdoor_VoxFomer} proposes a class-agnostic query proposal and class-specific segmentation for SSC. HASSC~\cite{ssc_outdoor_HASSC} introduces a hardness-aware approach that dynamically selects hard voxels based on global hardness and incorporates a self-distillation strategy to ensure stable and consistent training.

Existing methods adopt a single-stage framework for MSSC prediction but struggle to perform well in complex scenes. To address the above issue, we propose a novel framework to decompose MSSC into coarse MSSC and Masked Recurrent Network stages.
\section{Methodology}
\label{sec:method}
In this section, we first provide an overview of the problem formulation and our method (\cref{Overview}). Next, we give a brief introduction to the coarse MSSC stage (\cref{Coarse SSC}). Following this, we delve into the Masked Recurrent Network (\cref{Masked Recurrent Network}), which is the core of our approach. The masked recurrent network consists of an MS-GRU module (\cref{sec: MS-GRU}), a Distance Attention Projection (DAP) mechanism (\cref{sec: distance attention projection}), and a Mask Updating Module (\cref{sec: mask_updating}). Lastly, we introduce the loss function for model training (\cref{sec: loss function}).

\subsection{Overview} 
\label{Overview}
\noindent\textbf{Problem Formulation.} 
Given a single-view RGB image $I_{RGB}$, which is a partial observation of a 3D scene, monocular semantic scene completion aims to infer volumetric occupancy of the 3D scene, and each voxel is assigned with a semantic label $C_i$, where $i \in [0,1,..., N]$. $N$ is the number of semantic categories and $C_0$ represents the empty space. 
Our objective is to train a two-stage network $\{F_{1}, F_{2}\}$ with the learnable parameters $\theta_1$ and $\theta_2$. For the first stage, $h_0,m_{0}=F_{1}(I_\mathrm{RGB}; \theta_1)$ generate the coarse estimation feature $h_0$ and initial mask $m_0$. For the second stage, $\hat{Y}=F_{2}(h_0,m_0,x_0; \theta_2)$ takes initial estimation feature $h_0$, initial mask $m_0$, the feature of distance attention projection $x_0$ as input and generates the final completion results.

\noindent\textbf{Overall Architecture.} As shown in Fig.~\ref{fig:framework}, our method consists of two stages: coarse MSSC and Masked Recurrent Network. The coarse estimation stage takes RGB images as input, extracts 2D features through an encoder, and projects the 2D features into 3D space using the estimated depth values. Then, a 3D network performs the coarse estimation. The Masked Recurrent Network iteratively updates the coarse estimation, which consists of a set of MS-GRU, Mask Updating Module, and Distance Attention Projection. 
In each iteration, the MS-GRU updates the hidden state $h_{t}$ by processing the previous hidden state $h_{t-1}$, the previous mask $m_{t-1}$, and the distance attention projection feature.

\subsection{Coarse MSSC}
\label{Coarse SSC}
In this stage, we obtain the coarse MSSC. Given a single RGB input, we extract 2D features and estimate the depth map. The 2D features are projected into 3D space according to the depth map. Subsequently, the 3D features are fed into the 3D network to obtain the coarse MSSC results.

\noindent\textbf{2D Feature Learning.} 
We use the pre-trained ResNet-50~\cite{resnet} as our backbone to extract RGB features. Then we leverage the off-the-shelf depth prediction method~\cite{depth_Adabins} to predict the depth of each image pixel. 

\noindent\textbf{2D-3D Projection.}
To alleviate the gap between 2D and 3D, we apply surface projection~\cite{SATNet} to project the 2D features to corresponding 3D positions. Given the estimated depth image $I_d$, the intrinsic camera matrix $K \in \mathbb{R}^{3 \times 3}$ and extrinsic camera matrix $[R|t]\in \mathbb{R}^{3 \times 4}$, we map the 2D features to 3D positions according to equation $p_{u,v}= K[R|t]p_{x,y,z}$.

\noindent\textbf{3D Network for Coarse MSSC.} 
The projected 3D features are fed into the 3D network that is a stack of AIC~\cite{AICNet} modules, and we obtain the initial coarse MSSC results. The details of the 3D Coarse MSSC network are provided in the supplementary material.

\subsection{Masked Recurrent Network (MRN)}
\label{Masked Recurrent Network}
Masked Recurrent Network estimates a sequence of MSSC results $\{\hat{y}^{1},...\hat{y}^{N}\}$ and masks $\{\hat{m}^{1},...\hat{m}^{N-1}\}$ from the feature of initial coarse MSSC $h^0$, mask $m_0$ and the feature of distance attention projection $x_0$. With each iteration, the information of each MSSC prediction state is selectively memorized or forgotten by the proposed MS-GRU. MS-GRU is able to concentrate on occupied regions to boost performance by masks and exploit the sparsity of 3D scenes to alleviate the computation cost. We design the mask updating module to adjust the mask in each iteration. The proposed distance attention projection assigns different attention according to the distance to the observed surface, which alleviates the projection error and serves as the input for MS-GRU.

\subsubsection{Masked Sparse Gated Recurrent Unit}
\label{sec: MS-GRU}
Standard GRU operates on every position of the input features with dense convolutions, which is computationally inefficient and leads to high memory requirements, especially for processing 3D data. To address this concern, we propose the Masked Sparse Gated Recurrent Unit (MS-GRU), which enhances feature sparsity by applying a dynamic mask that focuses on the occupancy region and designs a sparse version of the GRU to leverage the sparsity for acceleration. Given the input feature $x_t$ of distance attention projection at the current time step, the hidden state $h_t$ is updated as:
\begin{align}
    z_{t}&=\mathrm{\sigma}(\mathrm{SubConv}([m_{t-1} \cdot h_{t-1},m_{t-1}\cdot x_{t}], W_z)),
    \\
    r_{t}&=\mathrm{\sigma}(\mathrm{SubConv}([m_{t-1} \cdot h_{t-1},m_{t-1}\cdot x_{t}], W_r)),
    \\
    h_{t}^{'}&=\mathrm{tanh}(\mathrm{SConv}([r_t \odot h_{t-1},m_{t-1}\cdot x_{t}], W_h)),
    \\
    h_{t}&=(1-z_{t}) \odot h_{t-1} + z_t \odot h_{t}^{'},
\end{align}
where $m_{t-1}$ represents the predicted binary mask of previous state, $z_{t}$ is the update gate, $r_{t}$ denotes the reset gate, $\mathrm{\sigma}$ indicates the sigmoid activation function, $\mathrm{tanh}$ is the hyperbolic tangent activation function. Moreover, $\mathrm{SubConv}$ and $\mathrm{SConv}$ represent the submanifold convolution and sparse convolution, respectively.
The proposed MS-GRU differs from the standard GRU in its mask and sparse design, with the specific details as follows.

\noindent\textbf{Mask Design}. In contrast to standard GRU, the mask $m_t$ is applied to the input $x_t$ and $h_{t-1}$, directing the GRU to focus only on the information within the mask. The mask $m_t$ indicates that if the location at $(x,y,z)$ is occupied. It can be defined as:
\begin{align}
    m(x,y,z)=\left\{
    \begin{array}{ll}
    1, & \mathrm{if\,~ voxel (x,y,z)\,~ is \, occupied~,}
    \\
    0, & \mathrm{if\,~ voxel (x,y,z)\,~ is \, empty~.}
    \\
    \end{array}
    \right.
\end{align}
The mask $m_t$ is updated by the Mask Updating Module (detailed in Sec.~\ref{sec: mask_updating}) in each iteration and supervised by the Sequential Mask Loss (detailed in Sec.~\ref{sec: loss function}).

\noindent\textbf{Sparse Design}. We leverage the mixture of submanifold sparse convolution~\cite{submanifold} and sparse convolution~\cite{sparse_Vote3deep} to design the MS-GRU block. Submanifold convolution and sparse convolution operate only on non-zero elements of the input tensor and their neighborhoods. Submanifold convolution preserves the sparsity in the output, while sparse convolutions can increase the sparsity of the output. When calculating the reset gate $z_t$ and update gate $r_t$, we apply submanifold convolution to ensure sparsity. In updating the hidden state $h_{t}^{'}$, we use sparse convolution to appropriately update the non-zero positions (occupied regions).

\begin{figure}
    \centering
    \includegraphics[width=\linewidth]{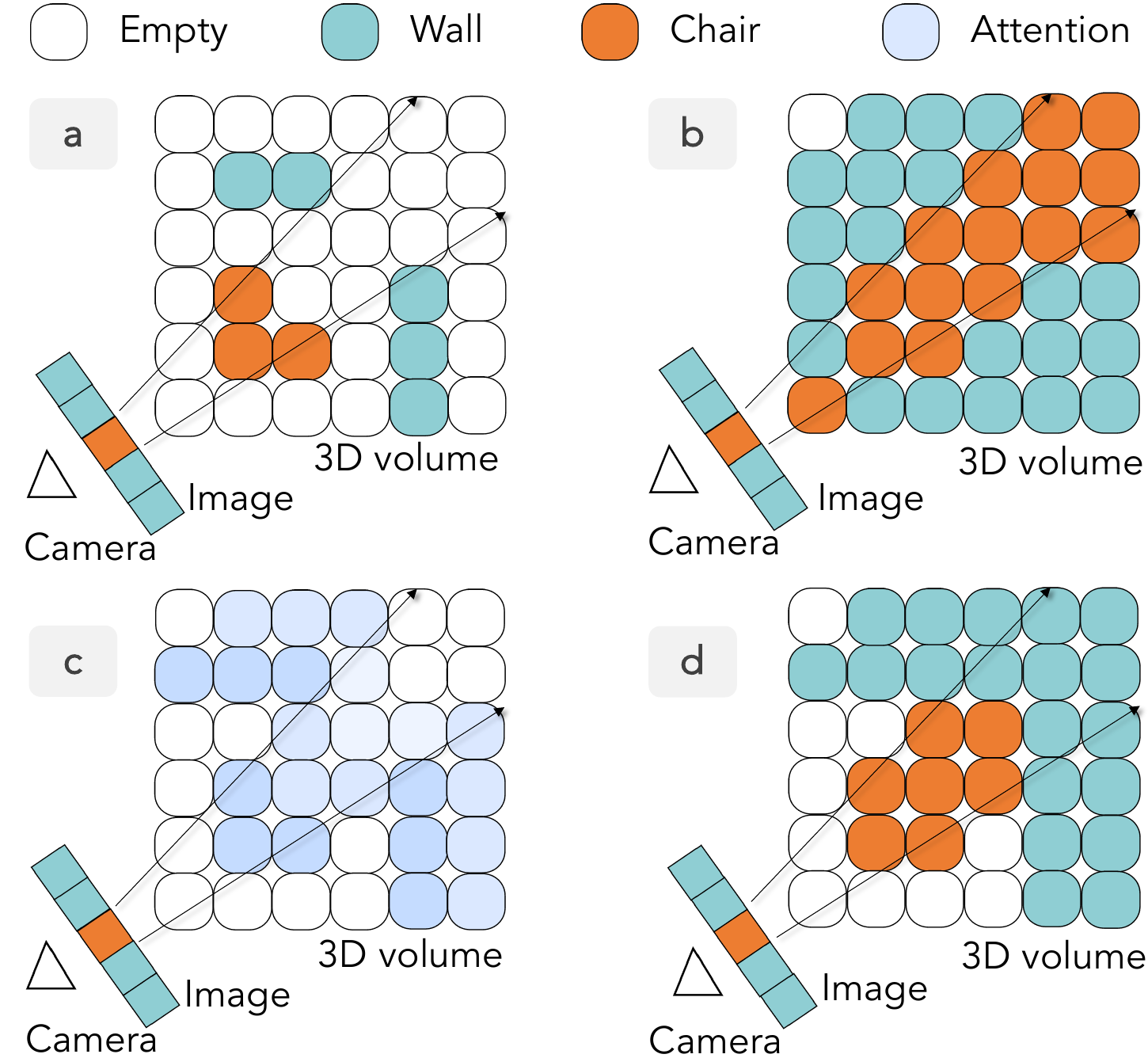}
    \vspace{-0.6cm}
    \caption{In (a) Surface Projection, voxels behind the surface are not assigned features. In (b) Sight Projection, although occluded regions are assigned features, a large number of incorrect features are also introduced. Distance Attention Projection determines the attention weight (c) based on the distance to the estimated surface and then weights the sight-projected features, allowing feature propagation to occluded regions while reducing the impact of introduced inaccurate features.}
    \label{fig:projections}
\end{figure}

\subsubsection{Distance Attention Projection}
\label{sec: distance attention projection}

At each iteration, \( x_t \) updates the hidden state \( h_t \). Ideally, \( x_t \) should be informationally complete, encapsulating essential data for updating \( h_t \), while maintaining high relevance and minimizing noise to ensure effective state transitions. We initially project the 2D RGB features into 3D space to mitigate the dimensional gap, and subsequently obtain $x_t$ through processing via convolutional layers. However, existing projection methods present several drawbacks as shown in Fig.~\ref{fig:projections}. Surface projection~\cite{SATNet,ssc_outdoor_VoxFomer} projects the 2D features onto a surface in 3D space based on the estimated depth. Consequently, voxels behind the surface are not assigned features, resulting in insufficient information to update the GRU in the occluded regions. Sight projection~\cite{ssc_outdoor_MonoScene}, on the other hand, maps 2D features to their corresponding line of sight without requiring depth data, allowing for rapid propagation of 2D features to distant occluded regions. However, it also introduces a significant number of incorrect features into these occluded regions.

To overcome these limitations, we introduce the distance attention projection, which assigns soft features to the 3D scene. This approach enables rapid feature propagation to occluded regions while allocating soft labels based on the correlation between 2D features and 3D regions. Consequently, it integrates the advantages of both surface and sight projections. The distance attention projection determines the attention weight $w_{d}$ based on the spatial occlusion relationship and the distance from different 3D locations to the estimated observation surface. The basic idea is that the further a point in space is from the estimated observation surface, the lower its relevance to the 2D image features, and thus the lower the attention weight $w_d$. 

We allocate attention weights based on the line of sight. In a 2D image, each pixel is part of a line of sight. The distribution of attention along the line of sight is described by the following formula:
\begin{align}
	w_{d}=\left\{
	\begin{array}{lr}
		1~/~(d-d'+1), &  d > d'\\
		1, & d = d'\\
		0.5, & \delta < d < d'\\
	\end{array}
	\right.
\end{align}
where $d'$ represents the distance from the camera to the observed surface, $d>d'$ denotes the occluded region in the line-of-sight, $d<d'$ indicates the region before the observed surface, and $\delta$ is defined according to the RMS of predicted depth.
The distance attention projection feature is obtained by element-wise multiplying the line-of-sight-projection with the attention matrix $w_d$. Subsequently, the feature of distance attention projection is fed into the AIC~\cite{AICNet} module to extract the 3D context feature $x_t$.

\begin{figure}
    \centering
    \includegraphics[width=\linewidth]{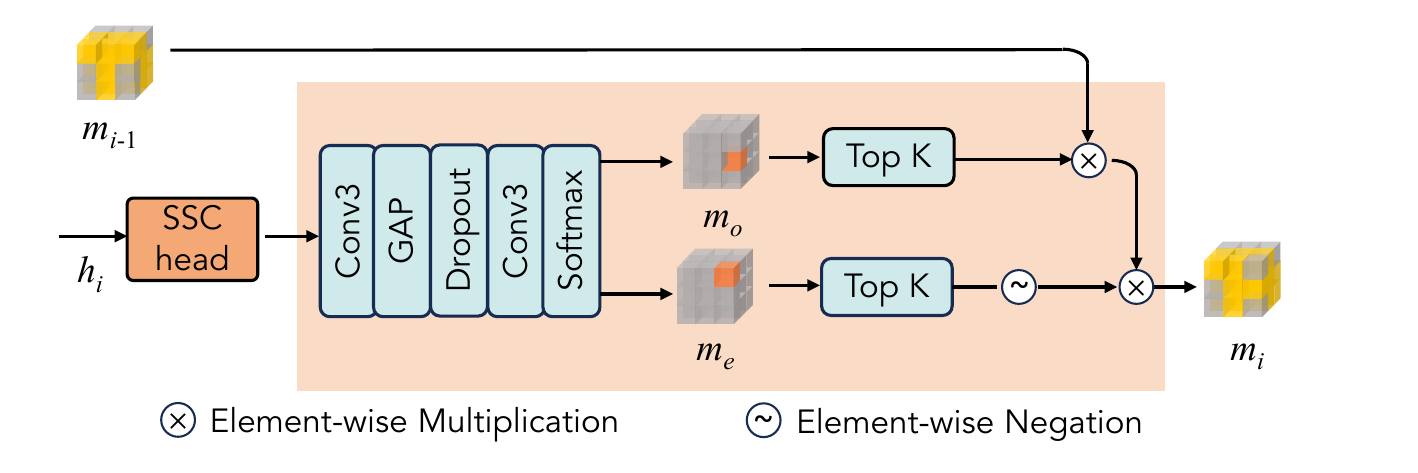}
    \caption{The architecture of Mask Updating Module. }
    \label{fig:mask head}
\end{figure}

\subsubsection{Mask Updating Module}
\label{sec: mask_updating}
The strong coupling between environmental occupancy patterns and object semantics has been established~\cite{SSCNet,ssc_outdoor_MonoScene}. We hypothesize that 3D scene occupancy information aids in voxel-wise semantic estimation. To leverage this, we integrate a mask into the standard GRU to emphasize occupied regions and introduce a mask updating module that iteratively enhances occupancy information. The mask \( m \) indicates whether a voxel is occupied or empty, supervised by the mask loss (see ~\cref{sec: loss function}). The ground truth for the mask is derived from the summation of all non-empty classes.

\noindent\textbf{Mask Initialization.} 
The mask $m$ is initialized during the Coarse MSSC stage and iteratively updated at each recurrent step. The initial mask $m_0$ is determined based on the occupancy score:
\begin{align}
	s_{x,y,z} = 1 - P_{x,y,z}(y=0 | I_\mathrm{RGB}; \theta_{1}),
\end{align} 
where $\theta_{1}$ denotes the Coarse MSSC network, $P(y=0)$ represents the probability to be empty and $1 - P(y=0)$ indicates the probability to be occupied. In practice, a predefined threshold $t$ is used: if $s_{x,y,z} \geq t$, then $m_0(x,y,z)$ is set to 1.

\noindent\textbf{Mask Updating Module.} 
We propose the mask updating module to sequentially update the mask $m$ as shown in \cref{fig:mask head}. The mask update module takes the output of the MSSC head as input, instead of the hidden state $h_t$ from the MS-GRU, in order to reduce the interference between the mask prediction and MSSC prediction tasks. The features are successively fed into a $3 \times 3$ convolution, GAP (Global Average Pooling) operation, dropout layer, $3 \times 3$ convolution layer, and softmax layer to obtain the probability of occupancy. The global average pooling operation is utilized to downsample the feature map two times. We use dropout to prevent overfitting by randomly setting $0.1$ of the input units to zero during training. After the softmax function, $m_o$ indicates the probability of occupancy for each voxel, and $m_e$ indicates the probability of emptiness for each voxel. Subsequently, we select the top $K$ voxels in $m_o$ and $m_e$. We add the top $K$ voxels in $m_o$ to $m_{i-1}$ and delete the top $K$ voxels in $m_{i-1}$ for updating.

\subsection{Objective \& Optimization}
\label{sec: loss function}

\noindent \textbf{Sequential MSSC Loss.} 
We first apply a sequential cross-entropy loss to both the initial coarse estimation $\hat y^{0}$ and each output state of recurrent refinement $\{\hat y^1, \hat y^2, ...,\hat y^N\}$, formulated as follows:
\begin{align}
    L_\mathrm{MSSC}=\sum_{i=0}^{N} \gamma^{i} L_\mathrm{ce}(y^\mathrm{gt},\hat y^{i}),
\end{align}
where $L_\mathrm{ce}$ indicates the cross-entropy loss and $\gamma=0.8$ in our experiments.

\noindent \textbf{Sequential Mask Loss.} 
The predicted masks from previous steps, \ie, $\{\hat m^1,\hat m^2,...,\hat m^N\}$, are supervised using a sequential binary weighted cross-entropy loss, defined as:
\begin{align}
    L_\mathrm{mask}=\sum_{i=1}^{N-1} \gamma^{N-i} L_\mathrm{wbce}(m_\mathrm{gt},m_{i}),
\end{align}
where $L_\mathrm{wbce}$ represents the binary weighted cross entropy loss and $\gamma=0.6$ in our experiments. Note that the ground truth for the mask is obtained by summing over all non-empty classes.

Overall, the total loss function can be formulated as:
\begin{align}
    \mathcal{L}_\mathrm{total} = \mathcal{L}_\mathrm{MSSC} +  \mathcal{L}_\mathrm{mask} + \mathcal{L}_\mathrm{scal},
\end{align}
where $L_\mathrm{scal}$ denotes the Scene-Class Affinity Loss proposed by MonoScene~\cite{ssc_outdoor_MonoScene}.
\definecolor{ceiling}{RGB}{214, 38, 40}
\definecolor{floor}{RGB}{43, 160, 43}
\definecolor{wall}{RGB}{158, 216, 229}
\definecolor{window}{RGB}{114, 158, 206}
\definecolor{chair}{RGB}{204, 204, 91}
\definecolor{bed}{RGB}{255, 186, 119}
\definecolor{sofa}{RGB}{147, 102, 188}
\definecolor{table}{RGB}{30, 119, 181}
\definecolor{tvs}{RGB}{188, 188, 33}
\definecolor{furniture}{RGB}{255, 127, 12}
\definecolor{objects}{RGB}{196, 175, 214}

\begin{table*}
    \centering
    \caption{
        \textbf{Comparisons among state-of-the-art MSSC methods} on the NYUv2 dataset. The symbol $\dag$ denotes the results presented by~\cite{ssc_outdoor_MonoScene}. The \hlorange{Best} and \hlteal{2nd Best} scores from each metric are highlighted in \hlorange{Orange} and \hlteal{Teal}, respectively.
    }
    \vspace{-0.2cm}
    \resizebox{\linewidth}{!}{
    \begin{tabular}{r|r|c|ccccccccccc|c|c}
    \toprule
    Method & Venue & \rotatebox{90}{SC IoU} & \rotatebox{90}{\textcolor{ceiling}{$\blacksquare$}~Ceiling} & \rotatebox{90}{\textcolor{floor}{$\blacksquare$}~Floor} & \rotatebox{90}{\textcolor{wall}{$\blacksquare$}~Wall} & \rotatebox{90}{\textcolor{window}{$\blacksquare$}~window} & \rotatebox{90}{\textcolor{chair}{$\blacksquare$}~chair} & \rotatebox{90}{\textcolor{bed}{$\blacksquare$}~bed} & \rotatebox{90}{\textcolor{sofa}{$\blacksquare$}~sofa} & \rotatebox{90}{\textcolor{table}{$\blacksquare$}~table} & \rotatebox{90}{\textcolor{tvs}{$\blacksquare$}~tvs} & \rotatebox{90}{\textcolor{furniture}{$\blacksquare$}~furniture} & \rotatebox{90}{\textcolor{objects}{$\blacksquare$}~objects} & \rotatebox{90}{SSC mIoU~} & FPS
    \\
    \midrule\midrule
    LMSCNet$^\dag$ \cite{ssc_outdoor_LMSCNet} & 3DV'20 & $33.93$ & $4.49$ & $88.41$ & $4.63$ & $0.25$ & $3.94$ & $32.03$ & $15.44$ & $6.57$ & $0.02$ & $14.51$ & $4.39$ & $15.88$&-
    \\
    AICNet$^\dag$ \cite{AICNet} & CVPR'20 & $30.03$ & $7.58$ & $82.97$ & $9.15$ & $0.05$ & $6.93$ & $35.87$ & $22.92$ & $11.11$ & $0.71$ & $15.90$ & $6.45$ & $18.15$& \hlorange{$3.68$}
    \\
    3DSketch$^\dag$ \cite{Sketch-Net} & CVPR'20 & $38.64$ & $8.53$ & $90.45$ & $9.94$ & $5.67$ & $10.64$ & $42.29$ & $29.21$ & $13.88$ & $9.38$ & $23.83$ & $8.19$ & $22.91$&\hlteal{$3.12$}
    \\
    MonoScene \cite{ssc_outdoor_MonoScene} & CVPR'22 & $42.51$ & $8.89$ & \hlteal{$93.50$} & $12.06$ & $12.57$ & $13.72$ & \hlteal{$48.19$} & \hlteal{$36.11$} & $15.13$ & $15.22$ & $27.96$ & $12.94$ & $26.94$ & $1.96$
    \\
    NDC-Scene \cite{ssc_NDC-scene} & ICCV'23 & \hlteal{$44.17$} & \hlteal{$12.02$} & \hlorange{$93.51$} & \hlteal{$13.11$} & $13.77$ & \hlteal{$15.83$} & \hlorange{$49.57$} & \hlorange{$39.87$} & \hlteal{$17.17$} & \hlorange{$24.57$} & \hlteal{$31.00$} & \hlteal{$14.96$} & \hlteal{$29.03$}&-
    \\ 
    \midrule
    \textbf{MonoMRN} & \textbf{Ours} & \hlorange{$53.16$} & \hlorange{$26.80$} & $92.02$ & \hlorange{$19.39$} & \hlorange{$18.50$} & \hlorange{$17.66$} & $44.60$ & $31.02$ & \hlorange{$19.60$} & \hlteal{$17.22$} & \hlorange{$32.90$} & \hlorange{$18.31$} & \hlorange{$30.73$} & $2.56$
    \\
    \bottomrule
    \end{tabular}}
    \label{tab:NYUv2}
\end{table*}
\begin{table*}
    \centering
    \caption{\textbf{Comparisons among state-of-the-art SSC methods} on the test set of SemanticKITTI \cite{dataset_semantickitti}. We include stereo (\textbf{S}), monocular (\textbf{M}) and temporal (\textbf{T}) methods for fair comparisons. `\textbf{I}', `\textbf{O}', and `\textbf{I/O}' denote methods that applied to indoor, outdoor, and both scenes. Symbol $\dag$ denotes the results presented in \cite{ssc_outdoor_MonoScene}. The \hlorange{Best} and \hlteal{2nd Best} scores from each metric are highlighted in \hlorange{Orange} and \hlteal{Teal}, respectively.}
    \vspace{-0.2cm}
    \resizebox{\linewidth}{!}{
    \begin{tabular}{r|c|c|c|ccccccccccccccccccc|c}
    \toprule
    Method & \rotatebox{90}{Input} & \rotatebox{90}{App. Scens.} & \rotatebox{90}{SC IoU} & \rotatebox{90}{\textcolor{car}{$\blacksquare$}~Car} & \rotatebox{90}{\textcolor{bicycle}{$\blacksquare$}~Bicycle} & \rotatebox{90}{\textcolor{motorcycle}{$\blacksquare$}~Motorcycle} & \rotatebox{90}{\textcolor{truck}{$\blacksquare$}~Truck} & \rotatebox{90}{\textcolor{other-vehicle}{$\blacksquare$}~Other-Vehicle} & \rotatebox{90}{\textcolor{person}{$\blacksquare$}~Person} & \rotatebox{90}{\textcolor{bicyclist}{$\blacksquare$}~Bicyclist}&\rotatebox{90}{\textcolor{motorcyclist}{$\blacksquare$}~Motorcyclist} & \rotatebox{90}{\textcolor{road}{$\blacksquare$}~Road} & \rotatebox{90}{\textcolor{parking}{$\blacksquare$}~Parking} & \rotatebox{90}{\textcolor{sidewalk}{$\blacksquare$}~Sidewalk} & \rotatebox{90}{\textcolor{other-ground}{$\blacksquare$}~Other-Ground~} & \rotatebox{90}{\textcolor{building}{$\blacksquare$}~Building} & \rotatebox{90}{\textcolor{fence}{$\blacksquare$}~Fence} & \rotatebox{90}{\textcolor{vegetation}{$\blacksquare$}~Vegetation} & \rotatebox{90}{\textcolor{trunk}{$\blacksquare$}~Trunk} & \rotatebox{90}{\textcolor{terrain}{$\blacksquare$}~Terrain} & \rotatebox{90}{\textcolor{pole}{$\blacksquare$}~Pole} & \rotatebox{90}{\textcolor{traffic-sign}{$\blacksquare$}~Traffic Sign} & \rotatebox{90}{SSC mIoU}
    \\
    \midrule\midrule
    \textcolor{gray}{VoxFormer-S ~\cite{ssc_outdoor_VoxFomer}} & \textcolor{gray}{S} & \textcolor{gray}{O} & $\textcolor{gray}{44.0}$ & \hlteal{$\textcolor{gray}{25.8}$} & $\textcolor{gray}{0.6}$ & $\textcolor{gray}{0.5}$ & $\textcolor{gray}{5.6}$ & $\textcolor{gray}{3.8}$ & \hlteal{$\textcolor{gray}{1.8}$} & \hlteal{$\textcolor{gray}{3.3}$} & $\textcolor{gray}{0.0}$ & $\textcolor{gray}{54.8}$ & $\textcolor{gray}{15.5}$ & $\textcolor{gray}{26.4}$ & $\textcolor{gray}{0.7}$ & $\textcolor{gray}{17.7}$ & $\textcolor{gray}{7.6}$ & $\textcolor{gray}{24.4}$ & $\textcolor{gray}{5.1}$ & $\textcolor{gray}{30.0}$ & $\textcolor{gray}{7.1}$ & $\textcolor{gray}{4.2}$ & $\textcolor{gray}{12.4}$
    \\
    \textcolor{gray}{HASSC-S ~\cite{ssc_outdoor_HASSC}} & \textcolor{gray}{S} & \textcolor{gray}{O} & \hlorange{$\textcolor{gray}{44.8}$} & \hlorange{$\textcolor{gray}{27.2}$} & $\textcolor{gray}{0.9}$ & $\textcolor{gray}{0.9}$ & \hlorange{$\textcolor{gray}{9.9}$} & \hlorange{$\textcolor{gray}{5.6}$} & \hlorange{$\textcolor{gray}{2.8}$} & \hlorange{$\textcolor{gray}{4.7}$} & $\textcolor{gray}{0.0}$ & \hlteal{$\textcolor{gray}{57.1}$} & $\textcolor{gray}{15.9}$ & $\textcolor{gray}{28.3}$ & $\textcolor{gray}{1.0}$ & $\textcolor{gray}{19.1}$ & $\textcolor{gray}{6.6}$ & \hlorange{$\textcolor{gray}{25.5}$} & $\textcolor{gray}{6.2}$ & \hlorange{$\textcolor{gray}{33.0}$} & \hlteal{$\textcolor{gray}{7.7}$} & $\textcolor{gray}{4.1}$ & $\textcolor{gray}{13.5}$
    \\
    \textcolor{gray}{H2GFormer-S \cite{ssc_outdoor_H2gformer}} & \textcolor{gray}{S} & \textcolor{gray}{O} & \hlteal{$\textcolor{gray}{44.2}$} & $\textcolor{gray}{23.4}$ & $\textcolor{gray}{0.8}$ & $\textcolor{gray}{0.9}$ & $\textcolor{gray}{4.8}$ & $\textcolor{gray}{4.1}$ & $\textcolor{gray}{1.2}$ & $\textcolor{gray}{2.5}$ & $\textcolor{gray}{0.1}$ & $\textcolor{gray}{56.4}$ & \hlteal{$\textcolor{gray}{26.5}$} & \hlteal{$\textcolor{gray}{28.6}$} & \textcolor{gray}{$4.9$} & \hlteal{$\textcolor{gray}{22.8}$} & $\textcolor{gray}{13.3}$ & $\textcolor{gray}{24.6}$ & \hlteal{$\textcolor{gray}{9.1}$} & $\textcolor{gray}{23.8}$ & $\textcolor{gray}{6.4}$ & $\textcolor{gray}{6.3}$ & \hlteal{$\textcolor{gray}{14.6}$}
    \\
     \textcolor{gray}{ MonoOcc-S \cite{ssc_outdoor_MonoOcc}} & \textcolor{gray}{M\&T} & \textcolor{gray}{O} & $\textcolor{gray}{-}$ & $\textcolor{gray}{23.2}$ & \hlorange{$\textcolor{gray}{2.2}$} & \hlteal{$\textcolor{gray}{1.5}$} & $\textcolor{gray}{5.2}$ & \hlteal{$\textcolor{gray}{5.4}$} & $\textcolor{gray}{1.7}$ & $\textcolor{gray}{2.0}$ & \hlteal{$\textcolor{gray}{0.2}$} & $\textcolor{gray}{55.2}$ & \textbf{$\textcolor{gray}{25.1}$} & $\textcolor{gray}{27.8}$ & \hlteal{$\textcolor{gray}{9.7}$} & $\textcolor{gray}{21.4}$ & \hlteal{$\textcolor{gray}{13.4}$} & $\textcolor{gray}{24.0}$ & $\textcolor{gray}{8.7}$ & $\textcolor{gray}{23.0}$ & $\textcolor{gray}{5.8}$ & \hlteal{$\textcolor{gray}{6.4}$} & $\textcolor{gray}{13.8}$
\\
      \textcolor{gray}{ HTCL-M \cite{ssc_outdoor_HTCL}} & \textcolor{gray}{S\&T} & \textcolor{gray}{O} & \hlteal{$\textcolor{gray}{44.2}$} & \hlorange{$\textcolor{gray}{27.2}$} & \hlteal{$\textcolor{gray}{1.8}$} & \hlorange{$\textcolor{gray}{2.2}$} & \hlteal{$\textcolor{gray}{5.7}$} & \hlteal{$\textcolor{gray}{5.4}$} & $\textcolor{gray}{1.1}$ & $\textcolor{gray}{3.1}$ & \hlorange{$\textcolor{gray}{0.9}$} & \hlorange{$\textcolor{gray}{64.4}$} & \hlorange{$\textcolor{gray}{33.8}$} & \hlorange{$\textcolor{gray}{34.8}$} & \hlorange{$\textcolor{gray}{12.4}$} & \hlorange{$\textcolor{gray}{25.9}$} & \hlorange{$\textcolor{gray}{21.1}$} & \hlteal{$\textcolor{gray}{25.3}$} & \hlorange{$\textcolor{gray}{10.8}$} & \hlteal{$\textcolor{gray}{31.2}$} & \hlorange{$\textcolor{gray}{9.0}$} & \hlorange{$\textcolor{gray}{8.3}$} & \hlorange{$\textcolor{gray}{17.1}$}
    
    \\
    \midrule
    LMSCNet$^\dag$~\cite{ssc_outdoor_LMSCNet} & M & O & $28.6$ & $18.3$ & $0.0$ & $0.0$ & $0.0$ & $0.0$ & $0.0$ & $0.0$ & $0.0$ & $40.7$ & $4.4$ & $18.2$ & $0.0$ & $10.3$ & $1.2$ & $13.7$ & $0.0$ & $20.5$ & $0.0$ & $0.0$ & $6.7$
    \\
    3DSketch$^\dag$~\cite{Sketch-Net} & M & I & $33.3$ & $18.6$ & $0.0$ & $0.0$ & $0.0$ & $0.0$ & $0.0$ & $0.0$ & $0.0$ & $41.3$ & $0.0$ & $21.6$ & $0.0$ & $14.8$ & $0.7$ & \hlteal{$19.1$} & $0.0$ & $26.4$ & $0.0$ & $0.0$ & $7.5$
    \\
    AICNet$^\dag$~\cite{AICNet} & M & I & $29.6$ & $14.7$ & $0.0$ & $0.0$ & $2.9$ & $0.0$ & $0.0$ & $0.0$ & $0.0$ & $43.6$ & $12.0$ & $20.6$ & $0.1$ & $12.9$ & $2.5$ & $15.4$ & $2.9$ & \hlteal{$28.7$} & $0.1$ & $0.0$ & $8.3$
    \\
    MonoScene~\cite{ssc_outdoor_MonoScene}& M & I/O & $34.2$ & $18.8$ & $0.5$ & $0.7$ & $3.3$ & \hlteal{$4.4$} & $1.0$ & $1.4$ & $0.4$ & $54.7$ & $24.8$ & $27.1$ & $5.7$ & $14.4$ & $11.1$ & $14.9$ & $2.4$ & $19.5$ & $3.3$ & $2.1$ & $11.1$
    \\
    VoxFormer-S~\cite{ssc_outdoor_VoxFomer} & M & O & \hlteal{$38.7$} & - & - & - & - & - & - & - & - & - & - & - & - & - & - & - & - & - & - & - & $10.7$
    \\
    TPVFormer~\cite{ssc_outdoor_TPVFormer} & M & O & $34.3$ & $19.2$ & $1.0$ & $0.5$ & \hlteal{$3.7$} & $2.3$ & $1.1$ & \hlteal{$2.4$} & $0.3 $& $55.1$ & \hlteal{$27.4$} & $27.2$ & \hlorange{$6.5$} & $14.8$ & $11.0$ & $12.9$ & $3.7$ & $20.4$ & $2.9$ & $1.5$ & $11.3$
    \\
    OccFormer~\cite{ssc_outdoor_OccFormer} & M & O & $34.5$ & $21.6$ & \hlteal{$1.5$} & \hlteal{$1.7$} & $1.2$ & $3.2$ & $2.2$ & $1.1$ & $0.2$ & $55.9$ & \hlorange{$31.5$} & \hlorange{$33.3$} & \hlorange{$6.5$} & \hlteal{$15.7$} & \hlteal{$11.9$} & $16.8$ & \hlteal{$3.9$} & $21.3$ & $3.8$ & \hlteal{$3.7$} & $12.3$
    \\
    NDC-Scene~\cite{ssc_NDC-scene} & M & I/O & $37.2$ & \hlorange{$26.3$} & \hlorange{$1.7$} & \hlorange{$2.4$} & \hlorange{$14.8$} & \hlorange{$7.7$} & \hlorange{$3.6$} & \hlorange{$2.7$} & $0.0$ & \hlteal{$59.2$} & $21.4$ & $28.2$ & $1.7$ & $14.9$ & $6.7$ & $19.1$ & $3.5$ & \hlorange{$31.0$} & \hlteal{$4.5$} & $2.7$ & \hlteal{$12.7$}
    \\
    \midrule
    \textbf{MonoMRN} & M & I/O & \hlorange{$42.0$} & \hlteal{$23.6$} & $0.8$ & $0.7$ & $2.6$ & \hlteal{$4.4$} & \hlteal{$2.4$} & $1.8$ & \hlorange{$0.4$} & \hlorange{$62.1$} & $22.5$ & \hlteal{$28.7$} & $5.6$ & \hlorange{$21.3$} & \hlorange{$16.0$} & \hlorange{$22.4$} & \hlorange{$5.1$} & $26.0$ & \hlorange{$7.9$} & \hlorange{$7.1$} & \hlorange{$13.8$}
    \\
    \bottomrule
    \end{tabular}
}
\label{tab:KITTI}
\vspace{-0.4cm}
\end{table*}

\section{Experiments}
We conduct an evaluation of our method using widely recognized real-world MSSC datasets, specifically the indoor NYUv2 \cite{dataset_nyu} and the outdoor SemanticKITTI \cite{dataset_semantickitti}. We compare our method with state-of-the-art MSSC methods. Then, we provide a detailed analysis of our evaluation results and the ablation studies. The qualitative performance of SemanticKITTI and more ablation studies are shown in the supplementary material.

\subsection{Experimental Setup}

\noindent\textbf{Datasets.} The NYUv2~\cite{dataset_nyu} dataset includes $1,449$ depth images captured by the Kinect from a diverse set of indoor scenes, partitioned into $795$ for training and $654$ for testing. Following SSCNet~\cite{SSCNet}, we use the 3D labels annotated by \cite{ssc15} and the mapping object categories based on \cite{ssc21}. 
SemanticKITTI \cite{dataset_semantickitti} is a large-scale autonomous driving dataset, which consists of $22$ point cloud sequences. Sequences $00$ to $10$, $08$, and $11$ to $21$ are used as training, validation, and testing, respectively. The LiDAR scans are represented as $256 \times 256 \times 32$ voxel grids of length $0.2$ meters. We use the RGB image of camera-2 and crop the image into size $1220 \times 370$.

\noindent\textbf{Evaluation Metrics.}
We follow the SSCNet \cite{SSCNet} protocol to ensure fair evaluations. For SSC, we use the intersection over union (IoU) between the predicted voxel labels and ground-truth labels for all object classes. The overall performance is evaluated by computing the mean intersection over union (mIoU) over all classes. To evaluate the scene completion performance, all voxels in the scene are categorized as empty or occupied. We compute IoU,  precision, and recall for scene completion.

\noindent\textbf{Implementation Details.}
We use the PyTorch framework to implement our method and conduct experiments. Our method adopts the SGD optimizer. We begin with an initial learning rate of $0.1$ and adjust it using the polynomial learning rate policy. We train the model for $200$ epochs on NYUv2 and for $30$ epochs on the SemanticKITTI dataset. The output 3D feature maps for NYU and SemanticKITTI are $60 \times 36 \times 60$ and $128 \times 128 \times 16$, respectively. For SemanticKITTI, the generated features are upsampled to $256 \times 256 \times 32$ to match the resolution of the GT for evaluation. MS-GRU is iterated $2$ times. The mask initialization threshold is set as $0.6$. We set the mask updating parameter $K$ to $5$.

\subsection{Comparisons with State of the Arts}
We compare our method with state-of-the-art methods, including those designed specifically for indoor scenes~\cite{AICNet,Sketch-Net}, outdoor scenes~\cite{ssc_outdoor_LMSCNet,ssc_outdoor_VoxFomer}, and those applicable to both indoor and outdoor scenes~\cite{ssc_outdoor_MonoScene,ssc_NDC-scene}. For outdoor MSSC, we also list the methods that use stereo images as input \cite{ssc_outdoor_VoxFomer,ssc_outdoor_HASSC,ssc_outdoor_H2gformer}. Stereo-based methods can provide more accurate geometric information. Taking VoxFormer \cite{ssc_outdoor_VoxFomer} as an example, it can achieve significant improvement when using stereo input as shown in \cref{tab:KITTI}.

\noindent\textbf{Quantitative Comparisons.} 
\cref{tab:NYUv2} and Table~\ref{tab:KITTI} report the performance of the proposed MonoMRN and other comparison methods on the indoor NYUv2 dataset (test set) and outdoor SemanticKITTI dataset (hidden test set). As shown in \cref{tab:NYUv2}, we significantly outperform the state-of-the-art methods on the NYUv2 dataset, achieving an $8.99\%$ IoU increase in SC and a $1.60\%$ mIoU improvement in SSC. As shown in Table~\ref{tab:KITTI}, our method outperforms state-of-the-art methods by $3.35\%$ in SC and by $1.07\%$ in SSC on the SemanticKITTI dataset.
Note that MonoScene and NDC-Scene adopt different 2D-to-3D projection strategies without using the estimated depth, which leads to significantly lower performance on the SC IoU compared with our methods. We additionally include comparison methods that utilize stereo and/or temporal images as input. While these methods leverage extra information to enhance performance, our approach still achieves competitive results without relying on such auxiliary data.

\begin{figure}
    \centering
    \includegraphics[width=\linewidth]{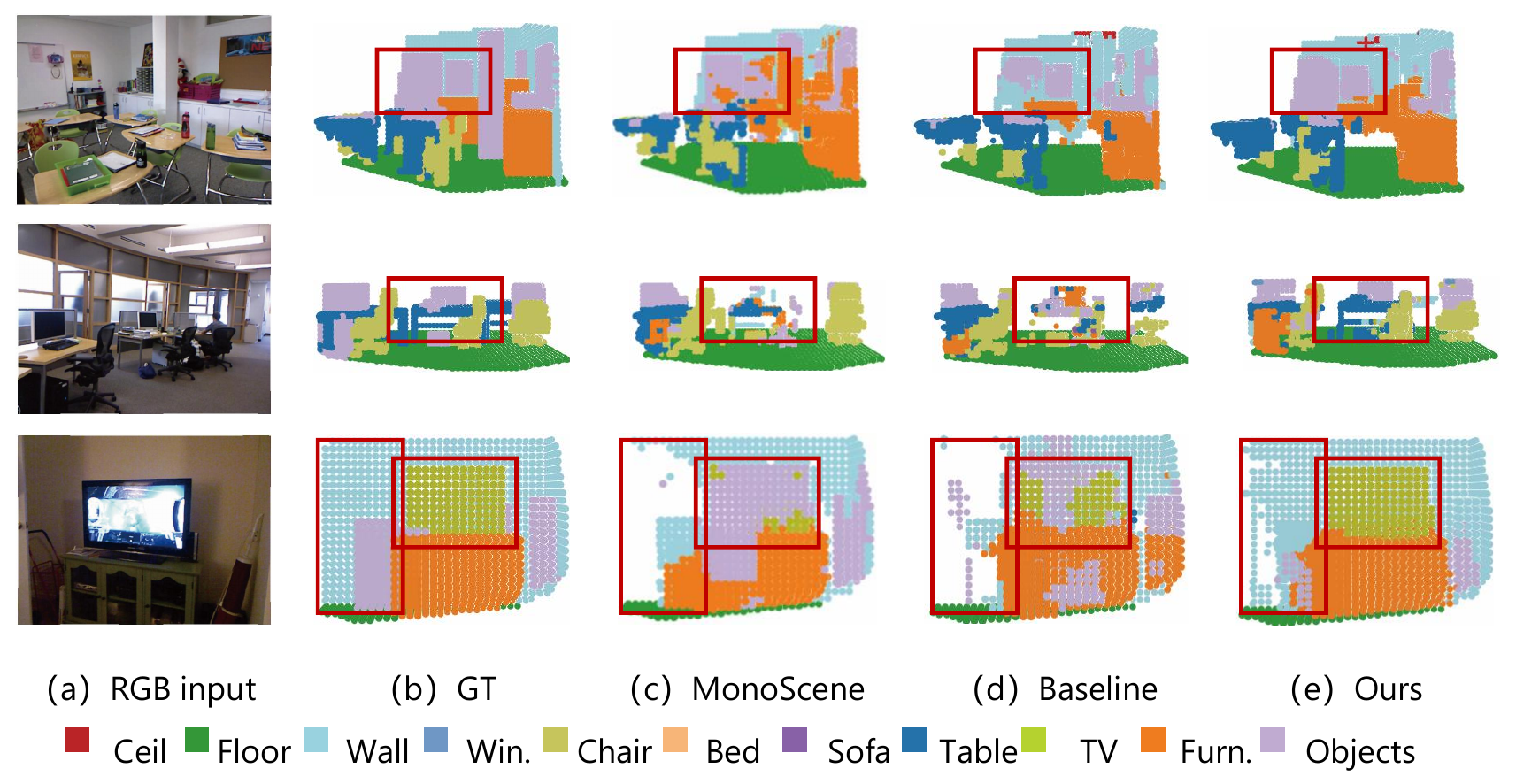}
    \caption{\textbf{Qualitative comparisons} of semantic scene completion results with different methods on the NYUv2 dataset. From left to right: (a) Single RGB input; (b) Ground truth; (c) MonoScene~\cite{ssc_outdoor_MonoScene}; (d) Our baseline method; and (e) Our approach.}
\label{fig:vis_nyu}
\end{figure}

\begin{figure}
    \centering
    \includegraphics[width=\linewidth]{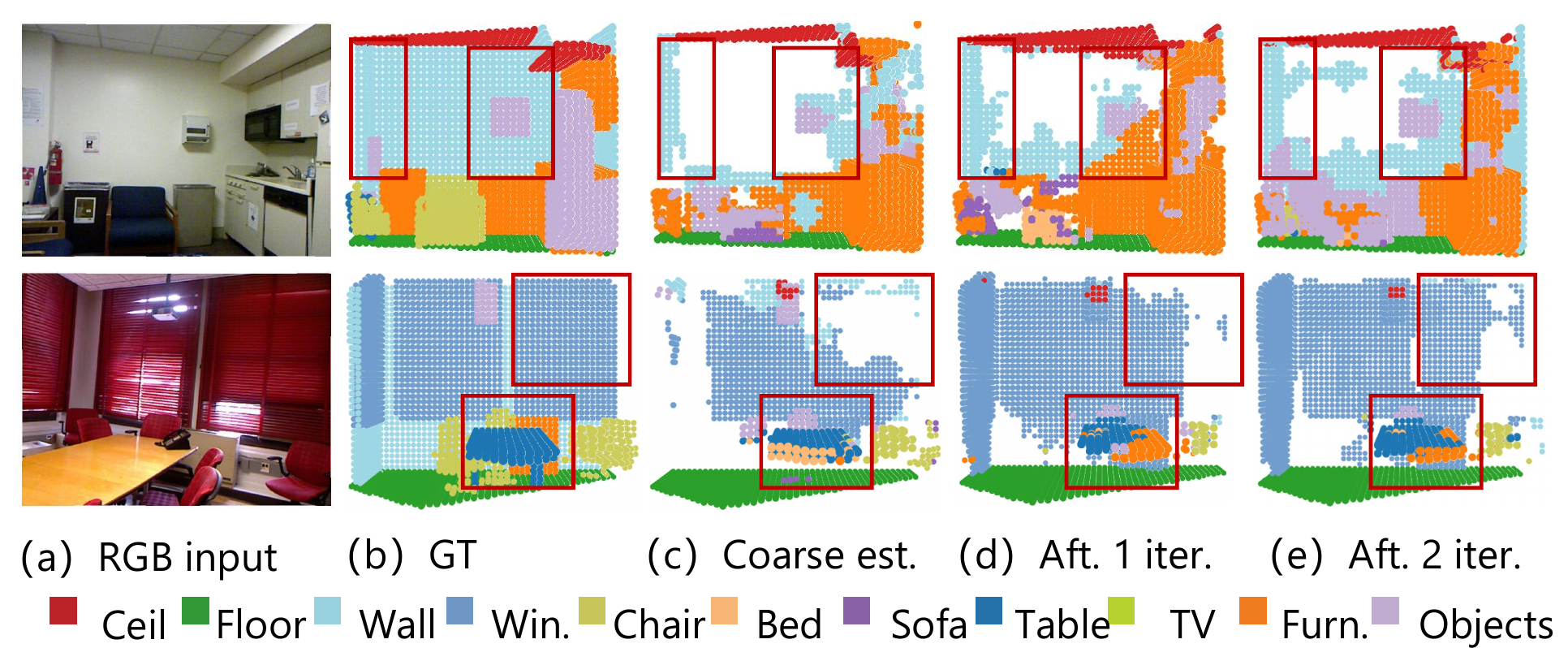}

    \caption{\textbf{Visual outputs} of MonoMRN at different stages. From left to right: (a) Single RGB input; (b) Ground truth; (c) Coarse estimation; (d) After one iteration; and (e) After two iterations.}
\label{fig:recurrent_vis}
\vspace{-0.2cm}
\end{figure}

\noindent\textbf{Qualitative Assessments.}
\cref{fig:vis_nyu} presents a comparison of the qualitative results on the NYUv2 dataset between our method, our baseline method, and MonoScene. We can observe that our method achieves the best qualitative results by restoring more precise details. By the proposed decomposed SSC framework, our method achieves noticeable improvement.  As NDC-Scene~\cite{ssc_NDC-scene} does not release the complete code, we can not compare our visualization results with it. Additionally, we visualized the outputs of MonoMRN at different stages. Fig.~\ref{fig:recurrent_vis} indicates that the Masked Recurrent Network can progressively refine the coarse MSSC results.

\subsection{Robustness Analysis}
In real-world scenarios, environmental disturbances are unavoidable. For instance, indoor MSSC systems often suffer from low lighting and motion blur, whereas outdoor MSSC systems are more vulnerable to adverse conditions such as fog and intense lighting. A model with high adaptability to such disturbances is crucial for practical applications~\cite{robo_ReliOcc,robo_robodepth,robo_robo3d,robo_pasco,robo_benchmark,xie2025robobev}. Therefore, in this section, we evaluate the improvement in model robustness achieved by the decoupled framework. Enhanced robustness not only enables the model to better handle complex environments but also reflects its improved capacity for information recovery \cite{xie2025drivebench,robo_robo3d,li2024is,hao2024is}, highlighting the role of the Masked Recurrent Network in the optimization process.
\begin{figure}
    \centering
    \includegraphics[width=\linewidth]{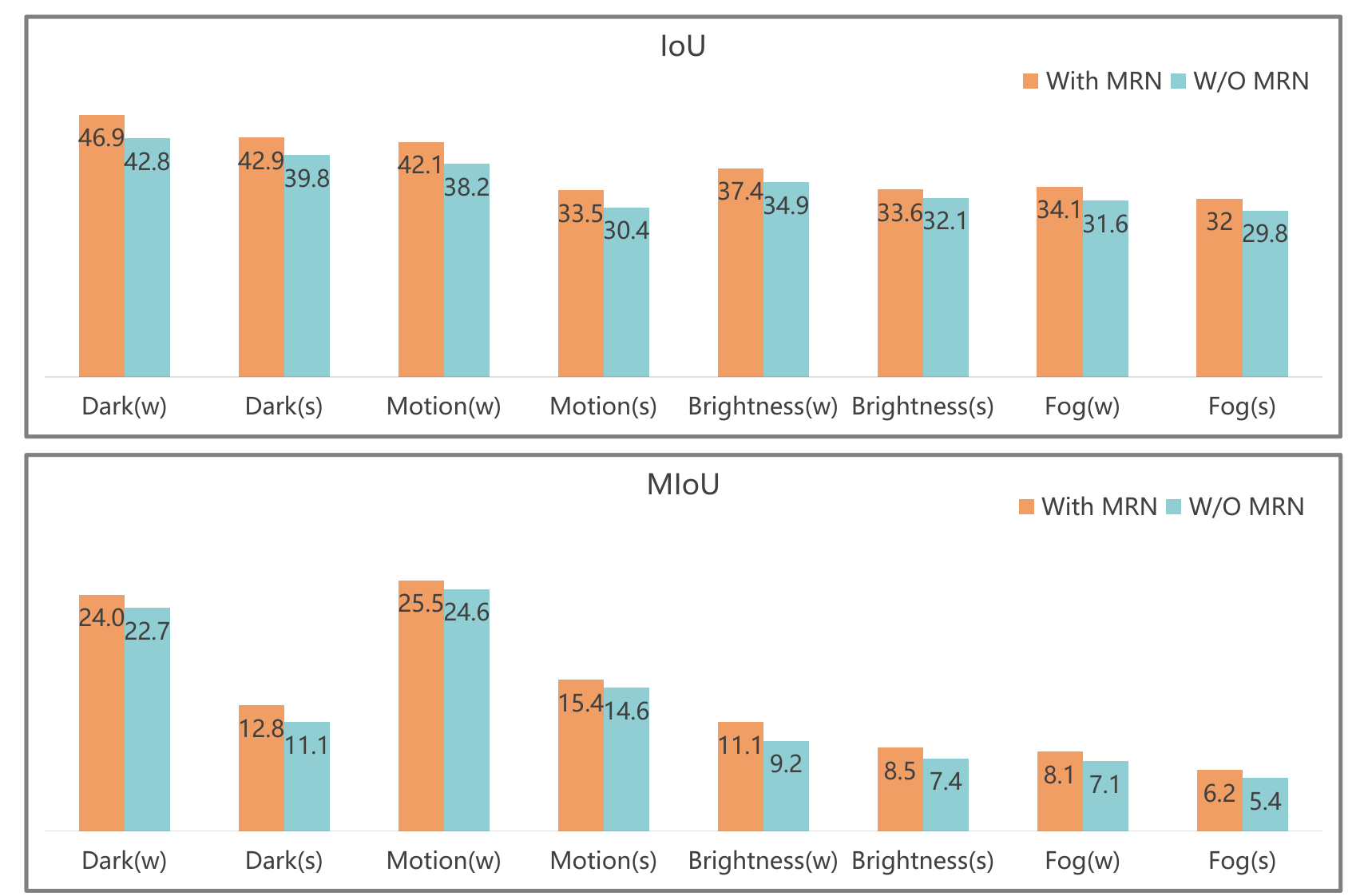}
    \caption{The robustness analysis with and w/o MRN Network. For indoor scenarios, we analyze robustness under dark and motion conditions, while for outdoor scenarios, we examine performance under brightness and fog conditions.}
    \label{fig:robustness}
    \vspace{-0.2cm}
\end{figure}

We simulate four potential out-of-distribution scenarios during inference: darkness and motion blur for indoor environments, and brightness and fog for outdoor environments. Each out-of-distribution scenario includes both weak (\textit{w}) and strong (\textit{s}) perturbation levels, providing a comprehensive evaluation of the model’s resilience to varying environmental disturbances. As shown in Fig.~\ref{fig:robustness}, the model exhibits enhanced robustness after processing through the Masked Recurrent Network under four types of disturbances, demonstrating stronger information recovery and completion capabilities.

\subsection{Ablation Study}
We present ablation results on the NYUv2 dataset to validate the effectiveness of our method from several perspectives: core components, Distance Attention Projection, the design choice of MS-GRU, and the design choice of mask updating. Due to space limits, more details and visualizations are placed in the Appendix.

\begin{table}
    \centering
    \caption{Ablation study on the core components in MonoMRN.}
    \vspace{-0.2cm}
    \resizebox{\linewidth}{!}{
    \begin{tabular}{l|c|c}
    \toprule
    \textbf{Method} & \textbf{SC-IoU} \textbf{(\%)} $\uparrow$ & \textbf{SSC-mIoU} \textbf{(\%)} $\uparrow$
    \\
    \midrule\midrule
    Baseline & $48.23$ & $27.47$          
    \\
    $+$ MS-GRU & $50.61$ $(+2.63)$ & $29.16$ $(+1.69)$          
    \\
    $+$ Distance Attention Projection & $51.86$ $(+1.25)$  & $30.11$ $(+0.95)$   
    \\
    $+$ Masking Updating & $\mathbf{53.16}$ $(+1.30)$ & $\mathbf{30.73}$ $(+0.62)$
    \\
    \bottomrule
    \end{tabular}}
    \label{tab:component_module}
    \vspace{-0.1cm}
\end{table}
\begin{table}
    \caption{Ablation study on the Distance Attention Projection.}
    \vspace{-0.2cm}
    \resizebox{\linewidth}{!}{
    \begin{tabular}{l|c|c}
    \toprule
    \textbf{Projection Method} & \textbf{SC-IoU} \textbf{(\%)} $\uparrow$ & \textbf{SSC-mIoU} \textbf{(\%)} $\uparrow$
    \\
    \midrule\midrule
    Surface Projection & $51.23$ & $29.36$          
    \\
    Sight Projection & $52.56$  & $30.12$          
    \\
    Distance Attention Projection&    $\mathbf{53.16}$  &$\mathbf{30.73}$ 
    \\
    \bottomrule
    \end{tabular}}
    \label{tab:distance_attention_projection}
    \vspace{-0.1cm}
\end{table}

\noindent\textbf{The Effectiveness of Core Components.}
We evaluate the effectiveness of the core components of our method, including MS-GRU, Distance Attention Projection, and Mask Updating, as shown in \cref{tab:component_module}. The vanilla baseline is the Coarse MSSC stage. The recurrent refinement by the proposed MS-GRU brings an improvement by $1.69 \%$ mIoU. The proposed Distance Attention Projection boosts the performance by $0.95\%$ mIoU. The proposed mask updating mechanism enhances the performance by $0.62\%$ mIoU.

\noindent\textbf{The Effectiveness of Distance Attention Projection.}
To validate the effectiveness of the proposed Distance Attention Projection, we replace it with Surface Projection and Sight Projection. As shown in \cref{tab:distance_attention_projection}, our method achieves the best performance among all comparisons. Compared with the other two methods, our approach efficiently assigns 2D features to occluded regions while reducing the impact of inaccurate assignments.

\noindent\textbf{Different Design Choices of MS-GRU.}
We evaluate the effectiveness of different design choices of MS-GRU as shown in Table~\ref{tab:abl_ms_gru}. We experiment with replacing the proposed MS-GRU with a standard GRU. We achieve better performance by using MS-GRU. We design a mixture of submanifold sparse convolution and sparse convolution in MS-GRU. Here, we test a version that only uses the submanifold sparse convolution. We achieve an improvement through the mixture design of sparse convolution.  We ablate a version of each MS-GRU to learn a separate set of weights. We can observe that when weights are tied, the performance is better and the parameter count is significantly lower. We conducted ablation experiments with different numbers of iterations. It can be observed that the performance increases with more iterations at first, then gets stable around $2$ and $3$. Due to the marginal improvement in the third iteration, we opt for two iterations in our method. Further increasing the number of iterations will degrade the performance.

\begin{table}
    \caption{Ablation study on different design choices of MS-GRU.}
    \vspace{-0.2cm}
    \resizebox{\linewidth}{!}{
    \begin{tabular}{c|c|c|c|c}
    \toprule
    \multirow{2}{*}{\textbf{Design Choice}} & \textbf{SC-IoU} & \textbf{SSC-mIoU} & \textbf{Params} & \textbf{MACs}
    \\
    & (\%) & (\%) & (M) & (G) 
    \\
    \midrule\midrule
    GRU & $50.48$ & $29.67$ & $1.33$ & $171.99$ 
    \\
    MS-GRU & $\mathbf{53.16}$ & $\mathbf{30.73}$ & $1.33$ & $\mathbf{52.44}$   
    \\
    \midrule
    Submanifold Conv & $52.66$ & $30.12$ & $1.33$ & $\mathbf{50.06}$     
    \\
    Sparse Conv & $\mathbf{53.16}$ & $\mathbf{30.73}$ &$1.33$ & $52.44$    
    \\
    \midrule
    Untied Weights & $52.85$ & $30.34$ & $2.66$ & $\mathbf{104.88}$   
    \\
    Tied Weights & $\mathbf{53.16}$ & $\mathbf{30.73}$ & $1.33$ & $\mathbf{104.88}$    
    \\
    \midrule
    $1\times$ Iteration & $53.01$ & $30.11$ & $1.33$ & $\mathbf{52.44}$  
    \\
    $2\times$ Iteration & $53.16$ & $30.73$ & $1.33$ & $104.88$  
    \\
    $3\times$ Iteration & $\mathbf{53.51}$&    $\mathbf{30.86}$ & $1.33$ & $157.32$   
    \\
    $4\times$ Iteration & $53.51$ & $30.17$ & $1.33$ & $209.76$   
    \\
    \bottomrule
    \end{tabular}}
\label{tab:abl_ms_gru}
\end{table}

\subsection{Limitations \& Discussions}
The predicted depth is far from the demand for semantic scene completion. For indoor scene completion, the depth estimation method \cite{depth_Adabins} obtains $0.364$ m
RMS on NYUv2, while the voxel size of our method is $0.08$ m. This indicates that the depth estimation algorithm still falls short of the requirements for MSSC and has a significant impact on the results. In the future, we will explore adding depth correction to our framework to further boost the performance.
\section{Conclusion}
In this paper, we proposed a novel decomposed framework that conducts Coarse MSSC followed by Masked Recurrent Networks. To better leverage the sparsity of 3D scenes and focus on the occupied 3D regions in each iteration, we introduced MS-GRU and the mask updating mechanism. Additionally, we proposed the Distance Attention Projection to mitigate projection errors. As a result, our approach achieves the state-of-the-art performance on the NYUv2 and SemanticKITTI datasets. We expect that the MonoMRN framework could inspire future research directions and drive more advancements in 3D perception tasks.

\section*{Acknowledgments}
This work was partially supported by the National Natural Science Foundation of China under Grant Nos. 62071330, 61831022, and U21B2020.
\\[0.5ex]
\noindent The authors would like to sincerely thank the Program Chairs, Area Chairs, and Reviewers for the time and effort devoted during the review process.
\vspace{0.1cm}

\section*{Appendix}
In this supplementary material, we first detail the architecture of the 3D network used in the initial MSSC. Next, we provide a qualitative comparison of the NYUv2 and SemanticKITTI datasets. We then include an ablation study on mask updating. 

\section{Details on the 3D network in Initial SSC}
\begin{figure}
\centering
\includegraphics[scale=0.5]{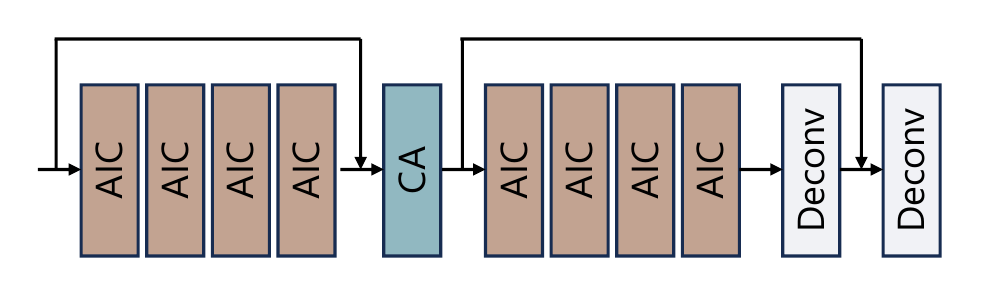}
\caption{Details on the 3D network in initial SSC.}
\label{fig:initial_SSC}
\end{figure}

Our method first performs an initial SSC prediction and then carries out the recurrent refinement. In this section, we detail the 3D network architecture in the initial SSC stage as shown in Fig.~\ref{fig:initial_SSC}. 

The projected 3D features are first passed through an encoder, which includes two AIC~\cite{AICNet} blocks and a channel-wise attention (CA) module~\cite{channel_attention1,channel_attention2}. Each block is composed of four AIC modules. An AIC module could model various objects or stuff with severe variations in shapes and layouts by an anisotropic receptive field. The channel-wise attention module is placed between the two AIC blocks to reweight and capture the channel-wise dependencies. After encoding, two deconvolution layers are applied to upsample the features to the original resolution of the input. Next, the features are fed into the SSC head to obtain the semantic scene completion results.
	
\section{Qualitative Comparisons}
\begin{figure*}
\centering
\includegraphics[scale=0.92]{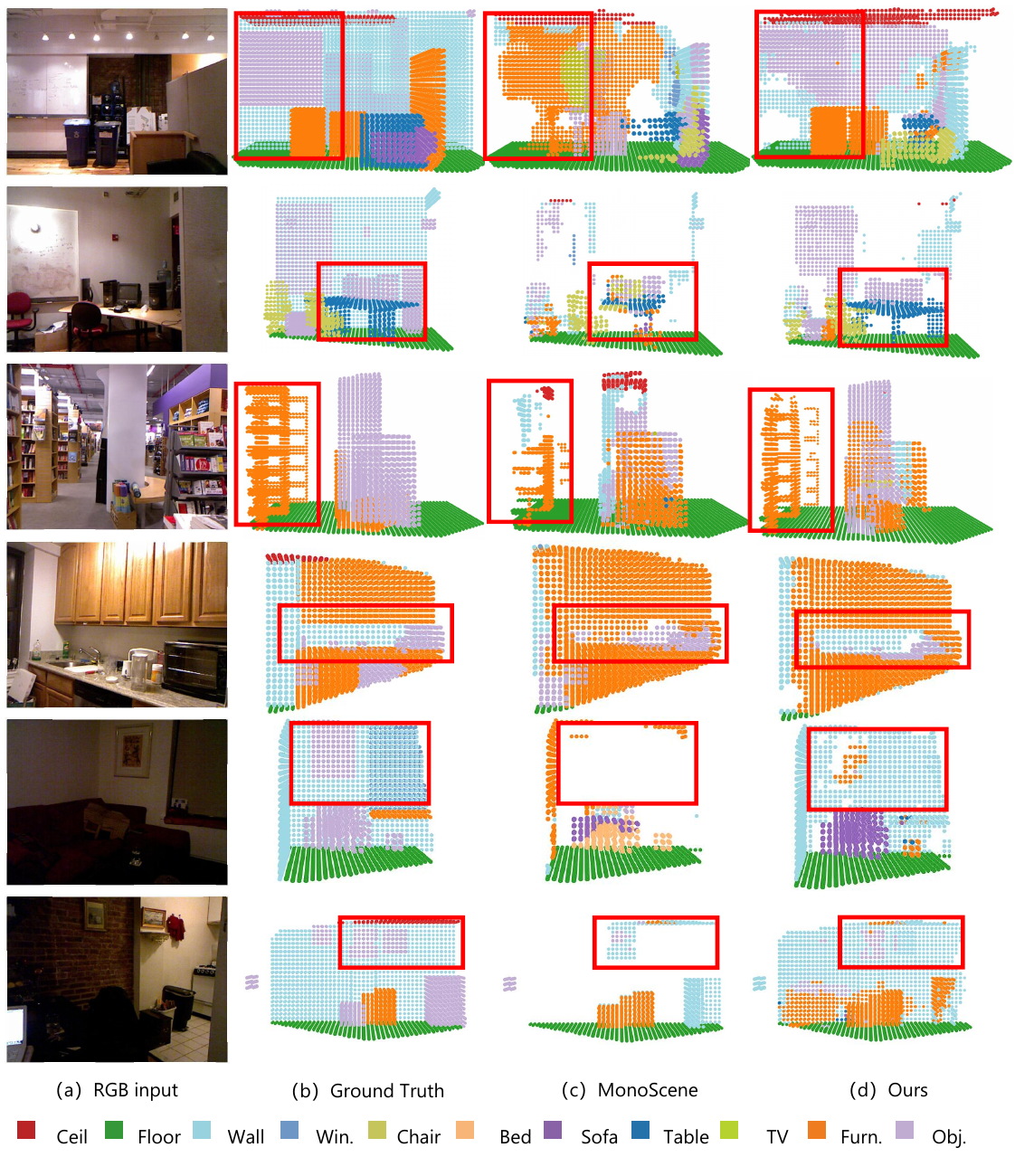}
\caption{Qualitative comparison on NYUv2. The leftmost column presents the input RGB images, while the subsequent columns sequentially show the results of Ground Truth,  MonoScene~\cite{ssc_outdoor_MonoScene}, and our method.}
\label{fig:supp_vis_nyu}
\end{figure*}

\begin{figure*}
\centering
\includegraphics[scale=0.52]{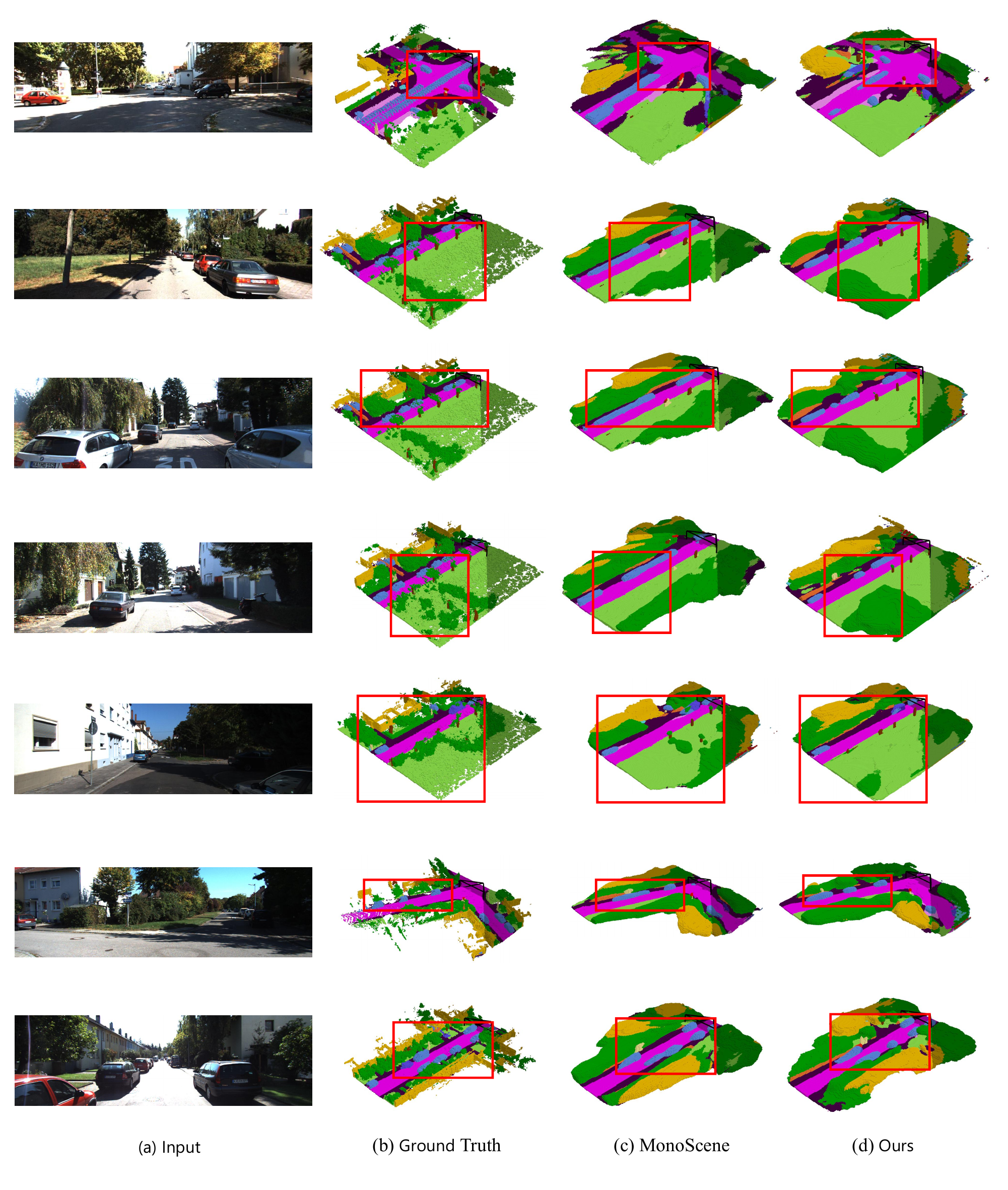}
\caption{Qualitative comparison on SemanticKITTI validation set. The leftmost column presents the input RGB images, while the subsequent columns sequentially show the results of Ground Truth, MonoScene~\cite{ssc_outdoor_MonoScene}, and our method.}
\label{fig:supp_vis_kitti}
\end{figure*}

Fig.~\ref{fig:supp_vis_nyu} and Fig.~\ref{fig:supp_vis_kitti} show the qualitative comparison on the NYUv2 and SemanticKITTI datasets. Our method outperforms MonoScene in the occluded regions and more effectively recovers fine-grained details. We choose MonoScene~\cite{ssc_outdoor_MonoScene} as a baseline for comparison because it targets the same type of scenarios as our method, providing a unified approach for both indoor and outdoor environments.
    
From Fig.~\ref {fig:supp_vis_nyu} and Fig.~\ref{fig:supp_vis_kitti}, it is evident that our method demonstrates superior capability in recovering various object categories, showcasing a stronger ability for information restoration. MonoScene struggles to accurately complete the semantic details of objects, often resulting in incomplete or imprecise reconstructions. In contrast, our method effectively captures finer details and restores more comprehensive semantic information, leading to a more accurate and visually coherent scene representation.

\section{Different Design Choices of Mask Updating}
In this section, we evaluate the effectiveness of different design choices for mask updating as shown in Table~\ref{tab:mask updating}.

\noindent\textbf{Mask Updating Module.} 
The mask updating module is proposed to sequentially update the mask $M$. We experiment with omitting the mask updating module. We can observe that it boosts $0.62\%$ mIoU by adding the mask updating module.

\noindent\textbf{Mask Initialization.} 
In the early stages of training, the mask predictions from the mask updating module are of low quality and fluctuate dramatically, causing the model to focus on inaccurate occupied regions. We tested a version that only used the mask updating module. Mask initialization could obtain $1.11\%$ performance gain.

\noindent\textbf{Mask Loss.} 
We introduce mask loss to provide supervision of the occupied regions. We can observe that mask loss enhances the performance by $1.06\%$ mIoU.

\begin{table}
\centering
\caption{Ablation study on the different design choices of the Mask Updating module in MonoMRN.}
\resizebox{\linewidth}{!}{
\begin{tabular}{r|c|c}
    \toprule[1pt]
    \textbf{Design Choices} & SC-IoU(\%) & SSC-mIoU(\%) 
    \\
    \midrule\midrule
    With Mask Updating Module & $52.33$ & $30.11$  
    \\
    W/O Mask Updating Module & $\mathbf{53.16}$ &   $\mathbf{30.73}$  
    \\
    \midrule[0.4pt]
    With Mask Initialization & $51.26$ & $29.62$     
    \\
    W/O Mask Initialization & $\mathbf{53.16}$ &     $\mathbf{30.73}$    
    \\
    \midrule[0.4pt]
    With Mask Loss & $51.66$ & $29.67$     
    \\
    W/O Mask Loss & $\mathbf{53.16}$ & $\mathbf{30.73}$    
    \\
    \bottomrule[1pt]
\end{tabular}}
\label{tab:mask updating}
\end{table}

\clearpage\clearpage
{
    \small
    \bibliographystyle{ieeenat_fullname}
    \bibliography{main}
}

\end{document}